\documentclass[11pt]{article}

\usepackage[final]{acl}

\usepackage{times}
\usepackage{latexsym}
\usepackage[T1]{fontenc}
\usepackage[utf8]{inputenc}
\usepackage{microtype}
\usepackage{inconsolata}
\usepackage{graphicx}
\usepackage{amsmath}
\usepackage{amssymb}
\usepackage{booktabs}
\usepackage{multirow}
\usepackage{xcolor} % to remove later
\usepackage[normalem]{ulem} % Required for strikethrough (\sout) to remove later

\title{Transformer Connectomics: Structural and Functional Connectivity\\
as a Network-Level Lens on Attention Heads}

\author{
  Daming Li\textsuperscript{1}\thanks{Corresponding authors: damingliyale22@gmail.com, jialu.zhang@uwaterloo.ca},
  Sophia Simeng Han\textsuperscript{2,3},
  Wanyu Lei\textsuperscript{4},
  Can Meng\textsuperscript{5},
  Jialu Zhang\textsuperscript{6}\footnotemark[1]
\\
  \textsuperscript{1}Independent Researcher,
  \textsuperscript{2}Stanford University,
  \textsuperscript{3}Zonaliz Inc.,
\\
   \textsuperscript{4}University of California, Berkeley,
  \textsuperscript{5}Yale University,
  \textsuperscript{6}University of Waterloo
}

\begin{document}
\maketitle
% Show page numbers (ACL style hides them by default)
\pagestyle{plain}

\begin{abstract}
Mechanistic interpretability has made substantial progress in identifying circuits responsible for specific behaviors in transformer language models, but offers comparatively few tools for characterizing the global organization of a model before a task or circuit is known. In this work, we adapt techniques from network neuroscience and introduce transformer connectomics, a network-level framework that applies the statistical methodology of structural and functional connectomics to attention heads. We treat attention heads as the nodes of a directed graph, define structural connectivity (SC) from weight-composition pathways and functional connectivity (FC) from correlations of head activity during inference, and analyze the resulting connectomes using permutation-based hypothesis tests, community detection, and structure--function coupling. We validate the framework along three complementary dimensions: recovering architectural organization, recovering known mechanistic structure, and distinguishing learned organization across models. Across these settings, transformer connectomics identifies meaningful head communities, captures established model structure without supervision, and reveals model-specific organizational patterns beyond architecture alone. Transformer connectomics complements existing circuit-discovery and feature-based approaches by providing a task-agnostic statistical description of head organization that can generate hypotheses for downstream mechanistic analysis. 

\end{abstract}

\section{Introduction}

Mechanistic interpretability (MI)~\citep{bereska2024mechanistic, sharkey2025open, somvanshi2026bridging} for transformer language models has matured rapidly. The Anthropic mathematical
framework~\citep{elhage2021mathematical} established attention heads as
discrete computational units writing to a shared residual stream,
\citet{olsson2022context} identified the induction circuit as the
substrate of in-context learning, \citet{wang2023interpretability}
mapped the indirect object identification (IOI) circuit in GPT-2 Small
to 26 specific heads, \citet{conmy2023towards} automated circuit
discovery via the ACDC algorithm, and dictionary-learning approaches
\citep{bricken2023monosemanticity,huben2024sparse} decompose
activations into sparse interpretable features.

Each of these tools has the same shape: \emph{find the circuit that
implements task $T$ in model $M$}. The starting point is a specific
behavior and the output is a small subgraph of heads (or features) causally
responsible for it. This is powerful but has missing pieces. Two questions are hard
to answer with task-specific tools:
(i) What does the population of heads look like as a whole, before
we know what task to ask about?
(ii) How do head--head relationships change across training,
architectures, or model families?

These are exactly the questions that network neuroscience~\citep{bassett2017network} answers for the organizational principle of brain regions \emph{at the system level}. Since the 1990s, neuroimaging has provided a global framework:
anatomical fiber tractography yields \emph{structural connectivity} (SC)~\citep{mori1999three, basser2000vivo, hagmann2008mapping} measuring the strength of anatomical connections between cortical areas, while
BOLD (Blood-Oxygen-Level-Dependent)-signal correlations across cortical regions yield \emph{functional
connectivity}
(FC)~\citep{biswal1995functional,friston2011functional} measuring the statistical synchronization of their neural activity over time.
These matrices are scored with statistical null distributions, clustered
into communities, summarized by graph-theoretic invariants, and compared
across subjects, ages, and species~\citep{sporns2005human, bullmore2009complex}. 

In this work, we adapt this statistical machinery to transformer-based large language models (LLMs), and introduce a framework called \emph{Transformer Connectomics} for analyzing the organization of attention heads. LLMs
are fundamentally different systems from real brains: deterministic trained weights and
inputs, free of biophysical constraints, optimized end-to-end on a well-defined next-token prediction objective. Therefore, the analogy is methodological only---we import
the analysis framework, not biological claims. 

Our contributions can be summarized as follows: (1) We establish a methodological analogy between LLMs and brains, and introduce the well-established notions in network neuroscience and neuroimaging to the MI studies of LLMs. (2) We define metrics and propose methods to measure the connection strengths, synchronization of activations at inference time, and the community structures in the network of attention heads. (3) We validate the framework by showing that it recovers architectural priors, known mechanistic circuits, and model-specific organizational differences across transformer families. Overall, by validating known structures unsupervised, we position our framework as a hypothesis generator for downstream causal and predictive analyses.

\section{Related Work}

\paragraph{Mechanistic interpretability.}
Mechanistic interpretability (MI) seeks to reverse-engineer the internal
representations of transformers to better understand the mechanisms and control the behaviors of LLMs~\citep{bereska2024mechanistic, sharkey2025open,somvanshi2026bridging}. The Anthropic mathematical
framework~\citep{elhage2021mathematical} cast attention heads as units
operating on a shared residual stream through QK/OV circuits and
identified induction heads as the substrate of in-context
learning~\citep{olsson2022context}. Circuit-level analyses then
localized specific behaviors to small subgraphs of heads---such as the
indirect-object-identification circuit in GPT-2
Small~\citep{wang2023interpretability}---and automated their discovery
via activation patching and ablation~\citep{conmy2023towards,
nanda2023progress}. A parallel line localizes and edits factual
knowledge stored in the feed-forward
layers~\citep{meng2022locating, geva2021transformer,
geva2023dissecting}. To address the polysemantic neurons that arise
from superposition~\citep{elhage2022toy}, dictionary learning
and sparse autoencoder methods decompose activations into interpretable
features at increasing
scale~\citep{bricken2023monosemanticity, huben2024sparse,
templeton2026scaling}.
Given a target behavior, these tools return
the minimal circuit or feature set responsible for it. 

Our framework is
complementary and task-agnostic, yielding a global head-level connectome
from which candidate circuits can be nominated. Closest to us, \citet{michel2019sixteen} and \citet{voita2019analyzing} performed head-level analyses and showed the strong heterogeneity in contributions of attention heads to model performance. 
\citet{sun2024brain} and \citet{kumar2024shared} directly fit LLM activations to human fMRI datasets to infer brain-like organization with LLMs. \citet{liu2026fnf} propose ICA-identified functional network fingerprints for model provenance prediction. Our framework operates at the attention-head level, is generic, and uniquely analyzes structure-function coupling, while claiming no biological correspondence.

\paragraph{Neuroimaging and network neuroscience.}
Network neuroscience models the brain as a graph whose nodes are
cortical regions and whose edges are structural or functional
connections~\citep{sporns2005human, bullmore2009complex,
bassett2017network}. Structural connectivity (SC) is estimated from
diffusion-MRI tractography~\citep{mori1999three, basser2000vivo, hagmann2008mapping} and functional
connectivity (FC) from correlations of resting-state BOLD (Blood-Oxygen-Level-Dependent)
fluctuations~\citep{biswal1995functional, friston2011functional}. These
connectomes are summarized with graph-theoretic
invariants~\citep{rubinov2010complex}, partitioned into intrinsic
functional networks and communities~\citep{power2011functional, raichle2001default}, and characterized by hub and rich-club
organization~\citep{van2011rich}. A key insight is
function relates to the underlying structure: resting-state FC is partly
predicted by SC~\citep{honey2009predicting}, and structure--function
coupling varies systematically along a cortical hierarchy~\citep{suarez2020linking, margulies2016situating}. We
introduce the statistical methods of permutation null models, community
detection~\citep{blondel2008fast}, and SC--FC coupling to
transformer-based LLMs, treating attention heads as the counterpart of cortical regions.

\section{Proposed Framework}
\label{sec:framework}

The brain is modeled as a graph in network neuroscience where structural connectivity reflects anatomical wiring—the white-matter pathways that physically carry signals between regions—while functional connectivity measures statistical dependence between regional activity, typically correlations of resting-state BOLD signals. We transfer this framework to transformer language models (Table~\ref{tab:analogy}), treating attention heads as nodes. The residual stream and inter-head composition are the counterpart of white matter, providing the channel through which one head's output influences another. Structural connectivity becomes a weight-only quantity measuring how strongly an earlier head writes into directions a later head reads, whereas functional connectivity is computed by correlating per-head activity over a large, diverse corpus, the task-agnostic analogue of a resting-state scan. As in neuroscience, structural and functional connectivity need not coincide, and comparing them through structure–function coupling is itself informative. This analogy is purely methodological: we do NOT claim that transformers work in the same way the brain does. Our implementation will be made available on GitHub \footnote{https://github.com/damingli09/transformer-connectomics}.

\begin{table}[h]
\small
\centering
\setlength{\tabcolsep}{4pt}
\resizebox{\columnwidth}{!}{%
\begin{tabular}{ll}
\toprule
\textbf{Brain} & \textbf{Transformer} \\
\midrule
Cortical region (parcel)    & Attention head \\
White-matter fiber pathway  & Residual stream + composition \\
Diffusion-MRI tractography  & $W_{OV}^A W_{QK}^B$ norms \\
BOLD signal correlation     & Per-head activity correlation \\
Resting-state scan          & Diverse corpus inference \\
\bottomrule
\end{tabular}}
\caption{Mapping between brain connectomics and LLM transformer connectomics introduced in this work. We adapt the analysis techniques, but do not make any biological claims for LLMs.}
\label{tab:analogy}
\end{table}

\subsection{Notation}

A transformer~\citep{vaswani2017attention} with $L$ layers and $H$
attention heads per layer has $N = L \cdot H$ head-nodes. Each node is
indexed by a flat integer $i = \ell \cdot H + h$, where $\ell$ is the
layer and $h$ the within-layer head index. Let $d$ denote the
residual-stream dimension and $d_h = d / H$ the per-head dimension. For
source $A = (\ell_A, h_A)$ and target $B = (\ell_B, h_B)$, we restrict
to forward-directed pairs ($\ell_A < \ell_B$): the residual stream gives
the connectome a strict directed acyclic structure. Each head $i$ carries
four learned weight matrices: the query, key, and value projections
$W_Q^i, W_K^i, W_V^i \in \mathbb{R}^{d \times d_h}$, which read the
residual stream into the head's $d_h$-dimensional query, key, and value
subspaces, and the output projection $W_O^i \in \mathbb{R}^{d_h \times d}$,
which writes the head's result back into the residual stream. Following
the circuits view of~\citet{elhage2021mathematical}, we combine these
into two per-head \emph{circuit matrices}: the output--value matrix
$W_{OV}^i = W_V^i \, W_O^i \in \mathbb{R}^{d \times d}$, which captures
\emph{what} the head writes to the residual stream, and the query--key
matrix $W_{QK}^i = W_Q^i \, (W_K^i)^{\top} \in \mathbb{R}^{d \times d}$,
which captures \emph{how} the head reads from it (its attention-pattern
map). Superscripts $A$ and $B$ denote these quantities for the source and
target heads, respectively.

\subsection{Structural Connectivity}

Intuitively, SC measures how strongly the output directions written by one head align with the subspaces read by another. Following~\citet{elhage2021mathematical}, we measure the structural
``bandwidth'' between heads $A$ and $B$ as the Frobenius norm of their
composition matrix. Two heads $A$ (at an earlier layer) and $B$ (at a later
layer) can compose via three pathways: $A$'s output enters $B$'s queries
(Q-composition), keys (K-composition), or values (V-composition):
\begin{align}
\mathrm{SC}_Q(A \!\to\! B) &= \| W_{OV}^A \, W_{QK}^B \|_F, \notag \\
\mathrm{SC}_K(A \!\to\! B) &= \| W_{OV}^A \, (W_{QK}^B)^\top \|_F, \\
\mathrm{SC}_V(A \!\to\! B) &= \| W_{OV}^A \, W_{OV}^B \|_F, \notag
\end{align}
with $\mathrm{SC}_{\text{total}} = \mathrm{SC}_Q + \mathrm{SC}_K
+ \mathrm{SC}_V$. For each model, the four SC matrices are strictly
upper-triangular in layer order and computed once from the model's
weights.

\paragraph{Low-rank computation.}
The naive evaluation materializes intermediates of shape $(L, H, d, d)$,
which for models such as Llama-3.1-8B with $d = 4096$ is $\approx 67$~GB. We exploit
the fact that both $W_{OV}$ and $W_{QK}$ are products of $(d, d_h)$ and
$(d_h, d)$ matrices and hence have rank at most $d_h$. By the cyclic
property of trace,
\begin{equation}
\| W_{OV}^A W_{QK}^B \|_F^2 = \mathrm{tr}\bigl( M^\top N_V[A] \, M \, N_K[B] \bigr),
\end{equation}
where $M = W_O^A W_Q^B \in \mathbb{R}^{d_h \times d_h}$,
$N_V[A] = (W_V^A)^\top W_V^A$, and $N_K[B] = (W_K^B)^\top W_K^B$, all
$d_h \times d_h$. Analogous identities hold for K- and V-composition.
Peak memory drops from $O(N d^2)$ to $O(N d_h^2)$, a $1000\times$
reduction for Llama-3.1-8B. The implementation matches the brute-force computation on GPT-2 Small to median per-entry relative error
$1.8 \times 10^{-5}$ while running $30\times$ faster.

\subsection{Activity Signal and Functional Connectivity}

The ``BOLD analog'' is the per-head residual-stream contribution. At
token $t$, head $i$ produces output $o_i(t) = z_i(t) W_O^i \in
\mathbb{R}^d$ where $z_i(t)$ is the pre-projection value-mixing output
(\texttt{hook\_z} in TransformerLens~\citep{nanda2022transformerlens}).
The scalar activity signal is $x_i(t) = \| o_i(t) \|_2$. Both FC
variants below are computed from these per-head scalar time courses.

We define two FC variants, each producing a symmetric $N \times N$
matrix with zero diagonal:

\noindent\textbf{L2}: z-scored Pearson correlation of $x_i(t)$ time
courses. This is the simplest variant that conflates magnitude with direction.

\noindent\textbf{Partial}: L2-norm Pearson correlation after regressing out the
residual-stream norm $r(t) = \| R^{(0)}(t) \|_2$ from each per-head
signal. This addresses the common-input confound: heads at the same
layer share input from the residual stream written by all earlier layers,
inflating within-layer FC by orders of magnitude.

Conceptually, L2 is the most direct BOLD analog: it correlates a single
scalar magnitude per head, asking only whether two heads are active on
the same tokens and thereby conflating how much a head fires with
in what direction it writes. Partial FC keeps the same scalar
signal but removes the shared residual-stream magnitude that every head
reads, analogous to the global-signal regression in
resting-state fMRI~\citep{fox2009global,murphy2009impact}. Because a transformer's defining feature is exactly
this shared communication channel, the common-input confound is the
dominant artifact in our setting. We therefore adopt partial FC as the
primary measure and report L2 alongside it as a simpler scalar baseline.

\subsection{Community Detection and Summary Statistics}

We build an undirected weighted graph from the positive entries of FC,
with adjacency $W_{ij} = \max(\mathrm{FC}_{ij}, 0)$ and $W_{ii} = 0$,
and apply Louvain modularity maximization~\citep{blondel2008fast}. The Louvain algorithm requires non-negative weights. We zeroed the negatives to avoid the artifact brought by global-signal regression, consistent with standard practices~\citep{rubinov2010complex}.
It greedily merges nodes to maximize the resolution-parameterized
modularity
\begin{equation}
Q(\gamma) = \frac{1}{2m} \sum_{i,j} \Bigl( W_{ij} - \gamma\,
\frac{k_i k_j}{2m} \Bigr)\, \delta(c_i, c_j),
\end{equation}
where $k_i = \sum_j W_{ij}$ is the weighted degree,
$2m = \sum_{ij} W_{ij}$, $c_i$ is the community assigned to node $i$,
and $\delta$ is the Kronecker delta. $\gamma$ is the resolution parameter of the modularity that the algorithm maximizes. Larger $\gamma$ favors more
smaller communities. We sweep $\gamma$ over $16$ values evenly spaced in
$[0.5, 2.0]$ with a fixed RNG seed and retain the partition with the
largest $Q(\gamma)$. We select the partition with the highest modularity because it is label free, to avoid leakage or circular statements.

We measure agreement between a discovered partition $\{c_i\}$ and a set
of published circuit labels $\mathcal{L} = \{(i, y_i)\}$, where $y_i$ is
the ground-truth class of labeled head $i$, by two metrics. \emph{Purity}
is the fraction of labeled heads that fall in their class's majority
community,
\begin{equation}
\mathrm{Purity} = \frac{1}{|\mathcal{L}|} \sum_{y}\,
\max_{c}\, \bigl| \{\, i : y_i = y,\ c_i = c \,\} \bigr|,
\end{equation}
and \emph{restricted} NMI is the normalized mutual information between
$\{c_i\}$ and $\{y_i\}$ evaluated only over the labeled subset,
\begin{equation}
\mathrm{NMI}_{\text{res}} = \frac{2\, I(C; Y)}{H(C) + H(Y)},
\end{equation}
where $I(C; Y)$ is their mutual information
and $H(\cdot)$ the Shannon entropy such that $\mathrm{NMI}_{\text{res}} \in
[0, 1]$. Restricting to the
labeled heads avoids assigning every unlabeled head a singleton
community, which would drown the signal in background entropy. Purity
measures how concentrated each ground-truth class is within a single
community, while restricted NMI additionally rewards separating distinct
classes; both are informative for multi-class label sets such as the
seven-class IOI circuit~\citep{wang2023interpretability}. For both
metrics, higher is better.

Four permutation-based nulls support hypothesis testing:
(i)~\emph{Phase-randomized FC null}: FFT-phase-shuffle each per-head
activity signal, preserving the power spectrum and marginal but
destroying cross-head temporal coupling.
(ii)~\emph{Configuration-model modularity null}: degree-preserving
double-edge swaps on the positive-weight FC graph, then re-apply Louvain.
(iii)~\emph{Weight-permutation SC null}: permute head indices within
each layer, then recompute SC.
(iv)~\emph{Within-label circuit-recovery null}: hold the discovered
partition fixed and permute the published circuit-class labels among the
labeled heads, then recompute NMI and purity; this tests whether the
communities align with the true class structure above chance.
All nulls use 10,000 resamples with the add-one $p$-value correction
of~\citet{phipson2010permutation}.

The principal summary statistic linking the two matrices is defined as:
\begin{equation}
\kappa = \mathrm{Pearson}\bigl( \mathrm{vec}_{\text{upper}}(\mathrm{SC}_{\text{total}}),\; \mathrm{vec}_{\text{upper}}(\mathrm{FC}) \bigr),
\end{equation}
restricted to entries with $\mathrm{SC} > 0$ (the directed-forward
edges). $\kappa \approx 1$ means FC mirrors SC capacity exactly;
$\kappa \approx 0$ means FC is decoupled from the structural pathway.
We also compute $\kappa$ within early, middle, and late layer thirds to
characterize the depth gradient.

\section{Validation I: Recovering Architectural Organization (GQA)}
\label{sec:caseE}

\subsection{Experimental Setup}

Original transformer models use standard multi-head attention~\citep{vaswani2017attention}. Modern
architectures such as Llama-3.1~\citep{grattafiori2024llama} adopt grouped-query attention
(GQA)~\citep{ainslie2023gqa}: multiple query heads share a single
$W_K, W_V$ projection. Llama-3.1-8B has $32$ query heads per layer but
only $8$ KV heads, so each KV head is shared by $G = 4$ query heads.
This gives $L = 32$, $H_q = 32$, $d = 4096$, $d_h = 128$, $N = 1024$
nodes. GQA is a clean structural prior with a testable prediction:
query heads in the same GQA group should be structurally and functionally
more similar to each other than to query heads in different groups.

We established three hypotheses before any Llama-3.1-8B data was
collected:

\begin{itemize}
\item \textbf{H1}: SC has a within-group block-diagonal signature. For
each forward-directed source--target pair $(\ell_A, \ell_B)$, the mean
cosine similarity of incoming-SC vectors among query heads in the same GQA group at
$\ell_B$ exceeds the mean cosine similarity across groups, by an exclusion-aware
group-label permutation null.
\item \textbf{H2}: Within-GQA-group FC magnitude exceeds between-group
FC magnitude, restricted to same-layer pairs.
\item \textbf{H3}: Louvain partitions on partial FC respect GQA group
boundaries: mean per-group fraction of query heads in the group's
majority Louvain community exceeds a within-layer shuffled null.
\end{itemize}

\paragraph{Implementation.}
The framework is built atop TransformerLens~\citep{nanda2022transformerlens} for model loading and hooks, PyTorch for tensor operations, NetworkX and python-louvain for
graph analysis, and scikit-learn for NMI. Code, data, and the
computed connectivity matrices will be released for community use. FC for $\textbf{H2}$ and $\textbf{H3}$ is computed from the first $50{,}000$ tokens of the SHA-pinned Pile corpus~\citep{gao2020pile} packed into $24$ sequences of length $2{,}048$, with the final partial sequence dropped. The choice of this corpus is for holding the input stimulus fixed across models in all the three results sections. On
Llama-3.1-8B ($N = 1024$), SC (low-rank) $+$ partial FC $+$ Louvain $+$
the H1/H2/H3 tests run in $\sim 15$~min on an H100 80\,GB GPU, with a
peak memory budget of $\sim 18$--$20$~GB.

\subsection{Results}

\begin{table}[h]
\small
\centering
\begin{tabular}{llcc}
\toprule
Test & Observed & Null $\mu \pm \sigma$ & Pass \\
\midrule
H1 (SC$_Q$) & $\Delta_{\cos}\!=\!+0.024$ & $0.000 \pm 0.0007$ & $\checkmark$ \\
H1 (SC$_K$) & $\Delta_{\cos}\!=\!+0.039$ & $0.000 \pm 0.0008$ & $\checkmark$ \\
H1 (SC$_V$) & $\Delta_{\cos}\!=\!+0.053$ & $0.000 \pm 0.0008$ & $\checkmark$ \\
H2          & $\Delta_{|FC|}\!=\!+0.184$  & $0.000 \pm 0.0034$ & $\checkmark$ \\
H3          & purity $= 0.795$            & $0.685 \pm 0.0067$ & $\checkmark$ \\
\bottomrule
\end{tabular}
\caption{Llama-3.1-8B GQA topology: all the pre-registered tests pass
at Bonferroni $\alpha = 0.01$. Permutation $p$-values hit the
$10{,}000$ resample resolution floor ($10^{-4}$).}
\label{tab:caseE}
\end{table}

\begin{figure*}[h]
\centering
\includegraphics[width=0.85\textwidth]{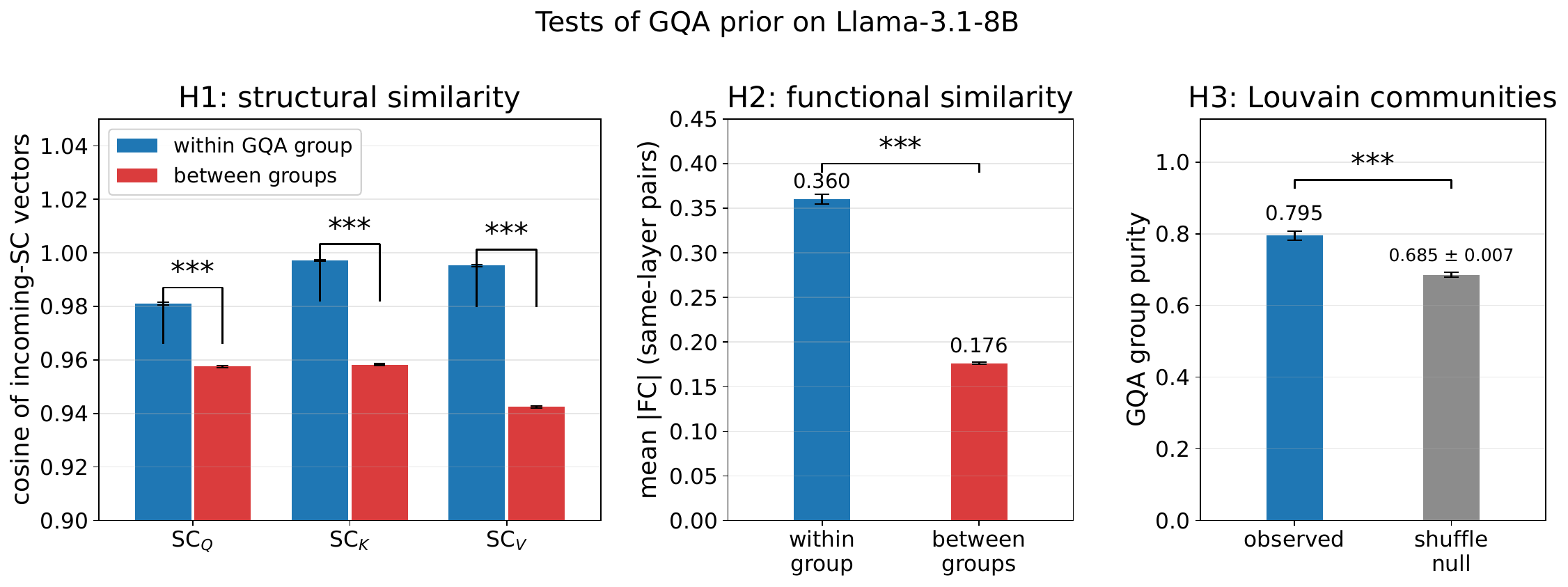}
\caption{Tests of GQA prior on Llama-3.1-8B. \textbf{Left:}
within-group vs.\ between-group SC cosine per composition pathway;
SC$_K$ and SC$_V$ are nearly identical within groups by construction.
\textbf{Middle:} FC magnitude is $2\times$ larger within GQA groups
than between. \textbf{Right:} unsupervised Louvain partitioning
respects GQA group boundaries far above a within-layer shuffle null.
Error bars show the Standard Error of the Mean (H1, H2) or the permutation-null SD (H3);
\textbf{***} denotes permutation $p$-value hitting the
$10{,}000$ resample resolution floor of $10^{-4}$.}
\label{fig:fig1}
\end{figure*}

All five tests (H1$\times$3 SC pathways + H2 + H3) pass at Bonferroni
$\alpha = 0.05/5 = 0.01$ with $10{,}000$ permutation nulls
(Table~\ref{tab:caseE}, Figure~\ref{fig:fig1}). The
structural-signature predictions (H1) are expected for
SC$_K$ and SC$_V$ by construction (within-group cosine $0.997$ and
$0.995$ respectively---heads sharing $W_K, W_V$ have nearly identical
incoming pathways), while a non-trivial within-group signature also
appears in SC$_Q$ (cosine $0.981$ within vs.\ $0.958$ between;
permutation $p$-values hit the
$10{,}000$ resample resolution floor of $10^{-4}$).

The functional-similarity prediction (\textbf{H2}) is another non-trivial
test: within-group FC magnitude ($0.360$) is $2.04\times$ the
between-group magnitude ($0.176$), permutation $p$-value hits the
$10{,}000$ resample resolution floor of $10^{-4}$. FC is computed from
activity, not weights. For it to reflect GQA structure, the shared
$W_K, W_V$ projections must induce correlated activation patterns at
the query heads they serve, which is not obvious---query heads share
the upstream projection but have their own $W_Q$ and $W_O$, so their
outputs could diverge.

The community-structure prediction (\textbf{H3}) is the clean
``framework discovers the prior'' test: unsupervised Louvain
partitioning recovers $79.5\%$ GQA group purity vs.\
$68.5 \pm 0.7\%$ null ($\sim 16\sigma$, permutation $p$-value hits the
$10{,}000$ resample resolution floor of $10^{-4}$). 

\begin{figure}[h]
\centering
\includegraphics[width=0.68\columnwidth]{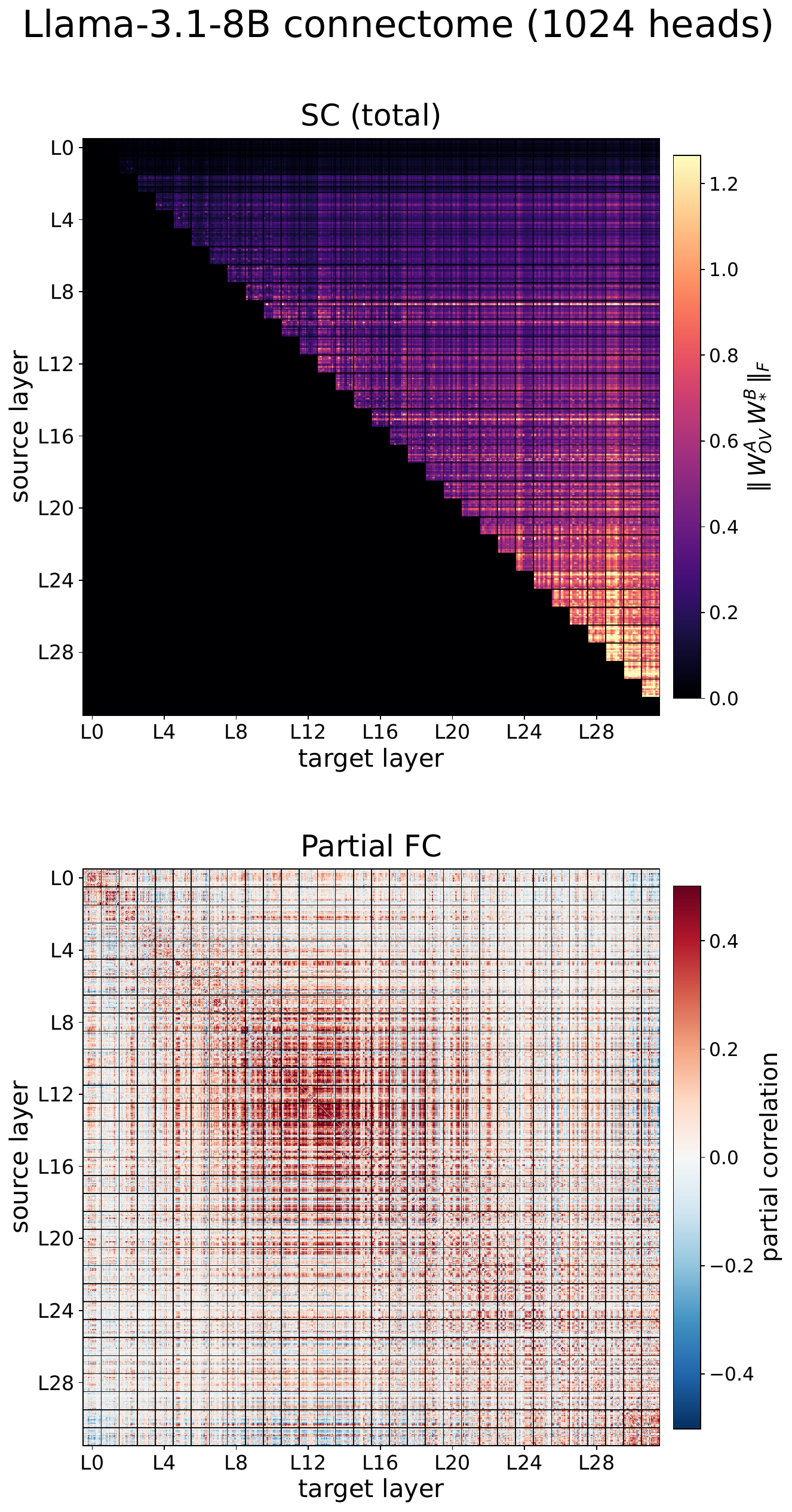}
\caption{Llama-3.1-8B SC (total) and partial FC on the $N = 1024$
head-node grid (32 layers $\times$ 32 query heads). SC is directed and
strictly upper-triangular. FC is symmetric. The depth-graded structure
in SC and the within-layer banding in FC are both visible at full
resolution.}
\label{fig:fig2}
\end{figure}

Figure~\ref{fig:fig2}
 shows the heatmaps visualizing the SC and partial FC matrices of Llama-3.1-8B.

\paragraph{$\kappa$ profile.}
Llama-3.1-8B's SC--FC$_{\text{partial}}$ coupling is
$\kappa_{\text{global}} = +0.075$,
$\kappa_{\text{early}} = +0.173$,
$\kappa_{\text{middle}} = +0.104$,
$\kappa_{\text{late}} = +0.011$: positive in early layers, near zero in
late. Llama-3.1-8B exhibits its strongest structure–function coupling in the earliest layers, with coupling weakening toward the top of the network. This suggests that functional interactions in early layers more closely follow the model's structural connectivity, whereas later-layer functional organization is less constrained by the static connectivity measure.

\section{Validation II: Recovering Known Mechanistic Structure (GPT-2 IOI circuit)}
\label{sec:validation}

\subsection{Experimental Setup}

We instantiate the framework on GPT-2 Small ($L\!=\!12$, $H\!=\!12$,
$d\!=\!768$, $N\!=\!144$~\citep{radford2019language}), loaded through
TransformerLens~\citep{nanda2022transformerlens}. We choose GPT-2 Small for this experiment, as there is a published circuit to compare against~\citep{wang2023interpretability}. The resting-state
corpus is a fixed $100{,}000$-token sample drawn with seed $0$ from the
\texttt{pile-10k} split of The Pile~\citep{gao2020pile}, tokenized with
the GPT-2 tokenizer and reshaped into non-overlapping length-$1024$
context windows. The corpus tensor is
SHA-pinned (\texttt{81ede231}$\ldots$) so every run---local or
cloud---sees byte-identical input. We run a single forward pass over the
corpus, cache each head's value-mixing output $z_i(t)$ via the
\texttt{hook\_z} hook, and form the activity signals
$x_i(t) = \lVert z_i(t)\, W_O^i \rVert_2$. From
these we compute the four SC matrices (weight-only) and both FC
variants, run the Louvain $\gamma$-sweep on the positive entries
of each FC, and score the resulting partitions against the published
$26$-head, seven-class IOI circuit
of~\citet{wang2023interpretability}.

The baseline is a random-weight architectural control: an
identically shaped GPT-2 Small with freshly re-initialized weights, pushed through the exact same pipeline on the same corpus. Because
architecture, corpus, and analysis are all held fixed, any difference
between the trained and random connectomes is attributable to the
learned weights alone. The end-to-end pipeline (SC $+$ both
FC variants $+$ Louvain sweep $+$ hypothesis tests) runs in $\sim 30$~min on a single A10 GPU.

\subsection{Results}

Table~\ref{tab:h1} shows that unsupervised Louvain partitioning partially recovers
the class structure of the IOI circuit~\citep{wang2023interpretability}
directly from ``resting-state activity''. Across both FC variants, restricted
NMI reaches $0.37$--$0.41$ and IOI purity reaches $0.81$--$0.85$, significantly higher than the random-weight control ($0.05$ NMI, $0.70$ purity), and the two
variants agree closely. All four tests are significant under a
within-label permutation null (IOI class labels shuffled among the labeled
heads, partition fixed; 10{,}000 resamples): NMI $p = 3\times10^{-4}$~(L2)
and $2\times10^{-3}$~(partial), purity $p = 3\times10^{-3}$~(L2) and
$6\times10^{-3}$~(partial), all below Bonferroni $\alpha = 0.05/4 =
0.0125$. The seven IOI roles are defined causally through
activation patching, whereas FC captures co-activation: that a purely
correlational, task-agnostic connectome already groups the IOI heads and
separates the trained circuit from the random baseline is exactly
what makes it a strong source of candidate circuits for downstream causal
analysis. Known mechanistic organization is thus reflected in the head
connectome and recovered without supervision. We visualize the GPT-2 Small
connectomes against a random-weight architectural control in
Appendix~\ref{sec:appendix-random}.

\begin{table}[t]
\small
\centering
\setlength{\tabcolsep}{5pt}
\begin{tabular}{l rr rr}
\toprule
& \multicolumn{2}{c}{IOI NMI} & \multicolumn{2}{c}{IOI purity} \\
\cmidrule(lr){2-3} \cmidrule(lr){4-5}
Measure & Trained & Random & Trained & Random \\
\midrule
L2      & 0.41 & 0.05 & 0.85 & 0.70 \\
Partial & 0.37 & 0.05 & 0.81 & 0.70 \\
\bottomrule
\end{tabular}
\caption{IOI-circuit recovery on GPT-2 Small. Unsupervised Louvain
partitioning partially recovers the seven-class IOI
circuit~\citep{wang2023interpretability} directly from resting-state
activity, achieving significantly higher restricted NMI and purity than the random-weight control. All four tests are significant under a within-label
permutation null (partition fixed, IOI class labels shuffled among the
labeled heads; 10{,}000 resamples, add-one $p$-value; null means NMI
$\approx0.17$, purity $\approx0.66$), each
below Bonferroni $\alpha=0.0125$.}
\label{tab:h1}
\end{table}

\section{Validation III: Distinguishing Learned Model Organization (GPT-2 vs.\ Pythia)}
\label{sec:crossmodel}

\subsection{Experimental Setup}

GPT-2 Small~\citep{radford2019language} and Pythia-160M~\citep{biderman2023pythia} are close in architecture sharing the same layer and dimension hyperparameters ($L=12$, $H=12$, $d=768$, $N=144$) but differ in
position embedding, and training procedures including corpus (WebText vs.\ the Pile), tokenizer, optimizer, and optimization schedule. This comparison asks whether transformer connectomics captures architectural structure alone or is sensitive to the learning recipes as well. 

To isolate the effect of learning, we drive both models with the identical SHA-pinned $100{,}000$-token corpus used in Section~\ref{sec:validation} and apply exactly the same analysis pipeline: computation of the four SC matrices, both FC variants, and Louvain community detection. Pythia-160M is evaluated at its final training checkpoint (step $143{,}000$). Because architecture shape, input corpus, and analysis pipeline are held fixed, any differences between the resulting connectomes arise from the learning process.

\subsection{Results}

\begin{figure}[t]
\centering
\includegraphics[width=\columnwidth]{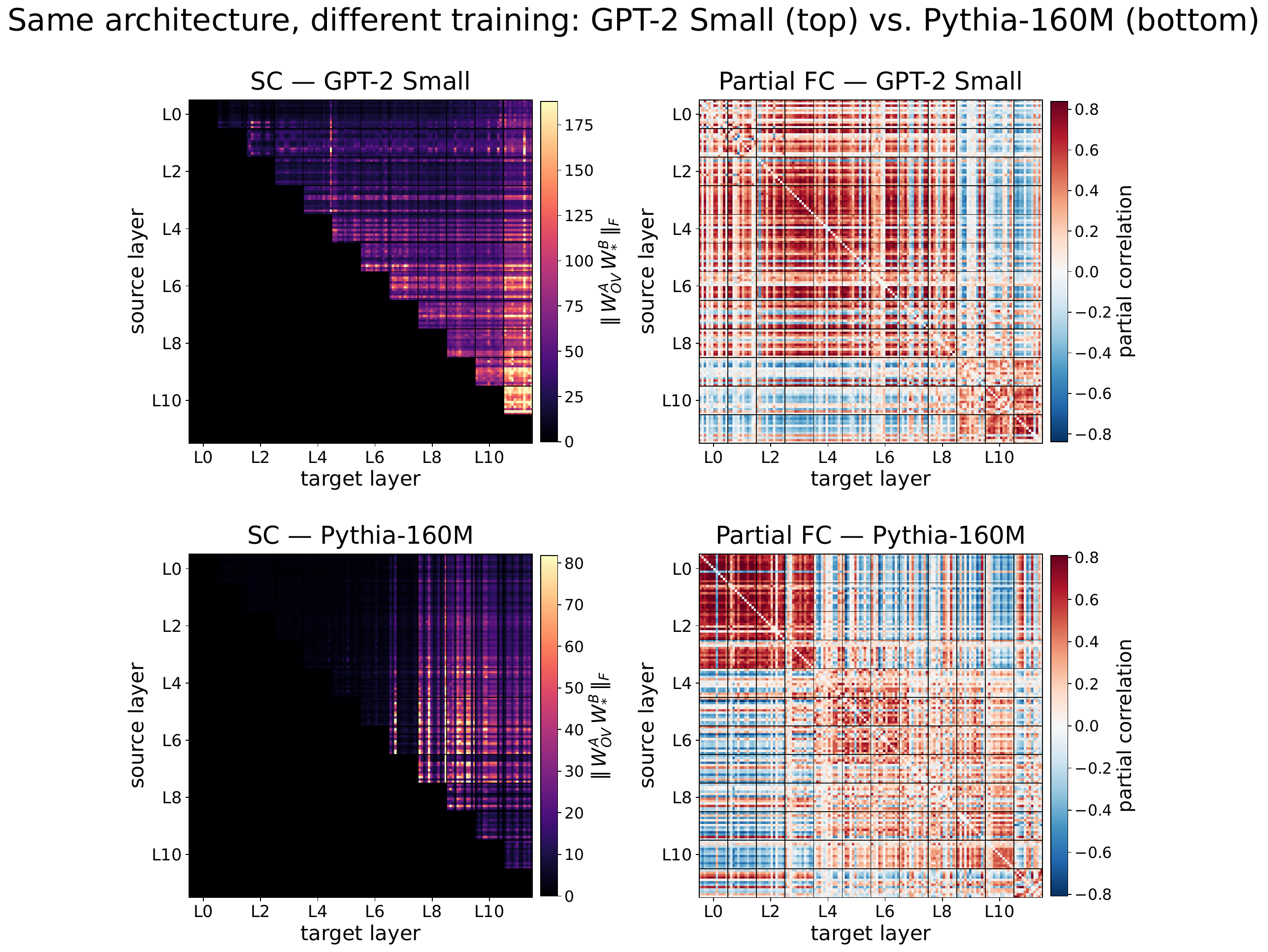}
\caption{GPT-2 Small (top) and Pythia-160M (bottom), both with 12L $\times$ 12H architecture. SC matrices differ in magnitude scale and
depth-graded structure; partial-FC matrices differ in within- vs.\
cross-layer balance.}
\label{fig:fig3}
\end{figure}

\paragraph{The connectome reflects learned organization, not architecture alone.} Figure~\ref{fig:fig3} compares the SC and partial-FC matrices for GPT-2 Small and Pythia-160M. Although the models share similar computational graphs, their connectomes are very different in both structural and functional organization. The differences extend beyond simple rescaling of edge weights: the depth-dependent structure of SC and the balance between within-layer and cross-layer FC are reorganized. Holding architecture shape, inference corpus, and analysis fixed therefore isolates the learning recipes as the primary source of these differences, indicating that transformer connectomics is sensitive to learned organization rather than architectural priors alone. This is in line with the concept of ``fingerprinting'' introduced by \citet{liu2026fnf}, indicating that the framework can be useful for various downstream use cases. Isolating the effects of individual training variables will be a future direction.

\paragraph{Structure-function coupling reveals different organizational profiles.} The distinction is particularly evident in the layer-resolved SC-FC coupling profile. GPT-2 Small exhibits its strongest SC-FC coupling in the middle and late layers ($\kappa_{\text{middle}} = +0.20$, $\kappa_{\text{late}} = +0.20$), whereas Pythia-160M concentrates coupling in the earliest layers ($\kappa_{\text{early}} = +0.40$) before falling to approximately zero or slightly negative values near the output ($\kappa_{\text{late}} = -0.07$). In contrast, the global coupling coefficient is close to zero for both models ($-0.03$ and $+0.00$), showing that a single scalar obscures substantial depth-dependent variation. The layerwise $\kappa$ profile therefore provides a more informative summary of structure-function relationships than the global statistic alone.

\paragraph{Shared architectural effects versus learned organization.} Both models exhibit strong within-layer FC, reflecting the shared residual stream that simultaneously provides input to all heads within a layer. This common-input effect is therefore a consequence of transformer architecture and appears consistently across models. In contrast, the detailed organization of SC, FC, and their layerwise coupling differs substantially despite identical architectures. The framework thus provides unique insights into individual models, instead of only reflecting architectural differences.

\section{Discussion}

MI has largely progressed through local analyses: given a behavior, identify the circuit or feature responsible for it. Our transformer connectomics approach addresses a complementary question by providing a global description of model organization before a specific task or circuit is known. Rather than replacing causal methods such as activation patching or circuit discovery~\citep{meng2022locating, wang2023interpretability}, it summarizes the statistical relationships among all attention heads using concepts from network neuroscience. This global view can reveal architectural regularities, characterize learned organization, and generate hypotheses for subsequent mechanistic investigation.

Across three validation settings, the framework consistently recovered meaningful organization at different levels. On Llama-3.1-8B it rediscovered the grouped-query attention prior without supervision, demonstrating sensitivity to architectural structure. On GPT-2 Small it partially recovered the class structure of the published IOI circuit from resting-state activity correlations alone, showing that known mechanistic organization is reflected in the connectome. Comparing GPT-2 Small and Pythia-160M further showed that models with nearly identical architectures nevertheless exhibit distinct connectomes, indicating that the framework captures learned computation beyond architecture alone.

More broadly, this work illustrates that connectomics can serve as an organizing layer for MI. SC, FC, community detection, and structure–function coupling are complementary views of the same network, providing a compact quantitative description of how computation is organized across the model.

The framework also opens several directions for future work. Community structure could be used to nominate candidate circuits for causal verification via activation patching or automated circuit-discovery methods. Applying the framework across training checkpoints may reveal how global organization emerges during optimization, while extending it across model scales and architectures could identify organizational principles that are conserved, or fundamentally different, across transformer families.

\section*{Limitations}

\textbf{Structural connectivity measures potential rather than realized communication.} Our SC metric is derived solely from learned weights and quantifies the capacity for one attention head to influence another through composition. It does not establish that the pathway is used for a particular input. Incorporating activation-dependent information into SC is a natural extension.

\textbf{Functional connectivity is correlational, not causal.} FC is computed from correlations of per-head activity and cannot distinguish direct interactions from shared upstream influences or indirect network effects. Partial FC reduces the dominant common-input confound from the residual stream, but causal relationships still require intervention-based methods. Future extensions can incorporate causality by adapting to the notions of effective connectivity \citep{friston2011functional} and dynamic causal modeling \citep{friston2003dynamic}.

\textbf{The framework currently models only attention heads.} MLP layers play an equally important role in transformer computation yet are excluded from the present connectome. Extending the graph to include MLP blocks or heterogeneous node types is an important direction for future work.

\textbf{The current evaluation focuses on validation rather than discovery.} We show that transformer connectomics recovers known architectural and mechanistic organization and distinguishes learned organization across models. While this is a vital prerequisite for a novel framework to establish the foundation necessary for future studies, demonstrating the discovery of previously unknown circuits or organizational principles will be natural next steps.

\bibliography{custom}

\appendix

\section{Trained vs.\ Random-Weight Control (GPT-2 Small)}
\label{sec:appendix-random}

\begin{figure}[h]
\centering
\includegraphics[width=\columnwidth]{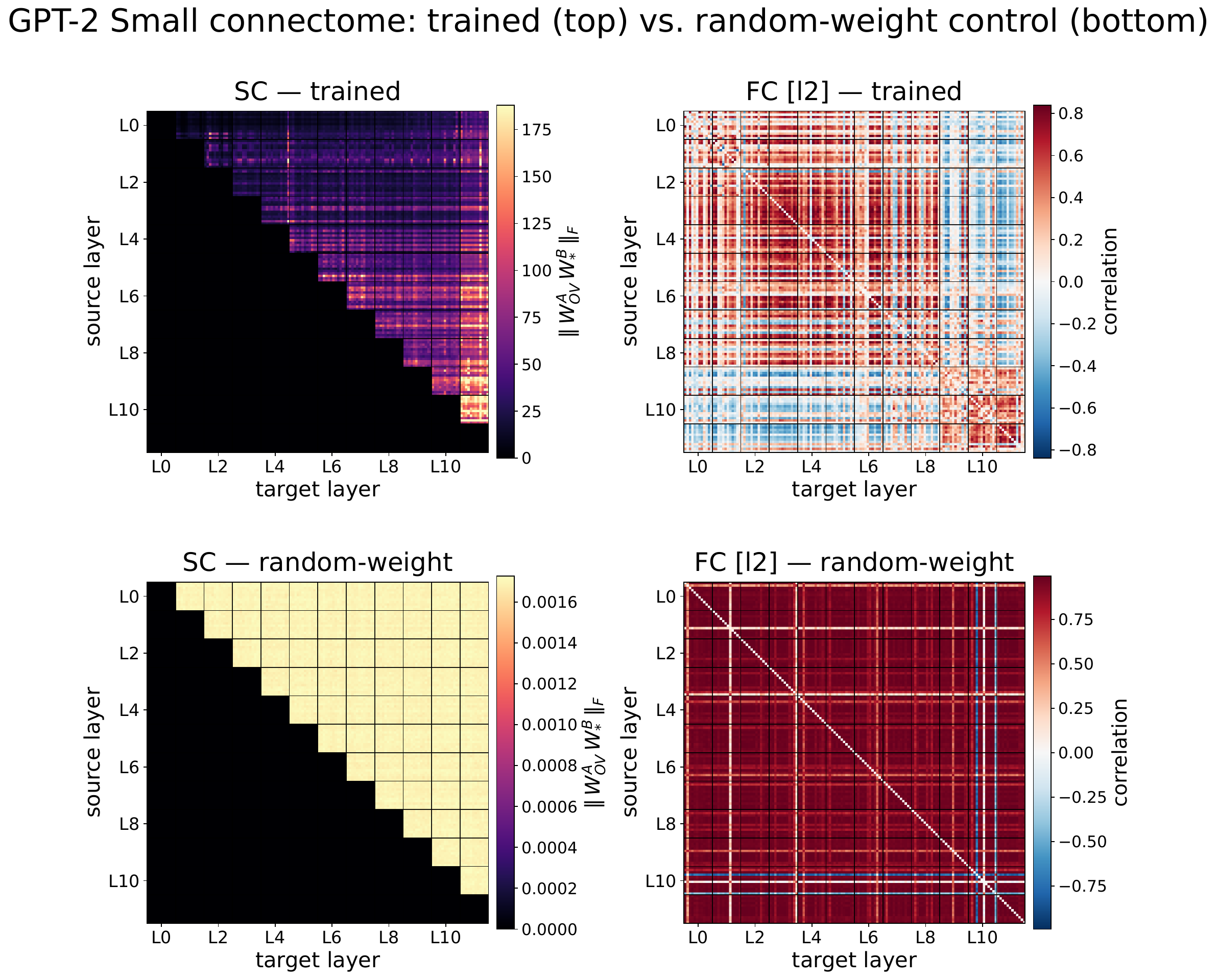}
\caption{Trained GPT-2 Small (top) vs.\ random-weight architectural
control (bottom), showing SC and FC matrices. Heads are ordered by
(layer, head). The trained model exhibits rich heterogeneous structure while the random model is uniform. Maximum SC differs by $10^5$, and modularity
$Q$ (partial FC) differs by $\sim$21$\times$.}
\label{fig:fig4}
\end{figure}

Beyond the circuit-recovery comparison in Section~\ref{sec:validation}, the
trained and random-weight connectomes differ starkly in their global
structure (Figure~\ref{fig:fig4}). Trained SC has rich heterogeneous
structure with depth-graded hub heads whereas random SC is uniform
low-amplitude. Modularity $Q$ on the primary FC measure (partial FC) is
$0.176$ for trained vs.\ $0.008$ for random ($\sim$21$\times$). The maximum SC entry
differs by five orders of magnitude. This confirms that the connectome
structure our framework analyzes is a property of the learned weights,
not of the architecture alone.

\end{document}